\documentclass[11pt,a4paper]{article}
\usepackage[hyperref]{acl2018}
\usepackage{times}
\usepackage{latexsym}
\usepackage{amsmath}
\usepackage{tikz}
\usepackage{tikz-dependency}
\usepackage[warn]{textcomp}
\usepackage[font=small]{caption}
\usepackage{subcaption}
\usepackage{multirow}
\usepackage{url}
\usepackage{etoolbox}
\usepackage{xr}
\newcommand{\com}[1]{}

\makeatletter
\patchcmd\@combinedblfloats{\box\@outputbox}{\unvbox\@outputbox}{}{%
   \errmessage{\noexpand\@combinedblfloats could not be patched}%
}%
 \makeatother

\usetikzlibrary{shapes,shapes.misc}

\aclfinalcopy % Uncomment this line for the final submission
\def\aclpaperid{600}

\title{Multitask Parsing Across Semantic Representations}

\author{Daniel Hershcovich$^{1,2}$ \\
  \\\And
  Omri Abend$^2$ \\
  $^1$The Edmond and Lily Safra Center for Brain Sciences \\
  $^2$School of Computer Science and Engineering \\
  Hebrew University of Jerusalem \\
  \texttt{\{danielh,oabend,arir\}@cs.huji.ac.il}
  \\\And
  Ari Rappoport$^2$
}

\date{}

\begin{document}

\maketitle

\begin{abstract}
  The ability to consolidate information of different types
  is at the core of intelligence, and has tremendous practical value
  in allowing learning for one task to benefit from generalizations learned for others.
  In this paper we
  tackle the challenging task of improving semantic parsing
  performance, taking UCCA
  parsing as a test case, 
  and AMR, SDP and Universal Dependencies (UD) parsing as auxiliary tasks.
  We experiment on three languages,
  using a uniform transition-based system and learning 
  architecture for all parsing tasks.
  Despite notable conceptual, formal and domain differences,
  we show that multitask learning significantly improves UCCA parsing
  in both in-domain and out-of-domain settings.
  Our code is publicly available.\footnote{\url{http://github.com/danielhers/tupa}}
\end{abstract}

\section{Introduction}\label{sec:introduction}

%The multitude of semantic representations put forth in recent years greatly enriches
%the discussion in NLP on the nature of semantic structure.
Semantic parsing has arguably yet to reach its full 
potential in terms of its contribution to downstream linguistic tasks,
partially due to the limited amount of semantically annotated training data.
This shortage is more pronounced in 
languages other than English, and less researched domains.

%While the development of syntactic treebanks has had a tremendous impact on natural 
%language processing, semantic representation has arguably yet to reach its full 
%potential in terms of its contribution to downstream linguistic tasks.
%As an example for syntactic annotation, the Universal Dependencies (UD) project provides
%cross-linguistically consistent treebanks in many languages \cite{nivre2016universal},
%and accurate parsers based upon it and other datasets have been extremely useful in natural
%language understanding tasks \cite{P16-1139,E17-1117,K17-3002}.
%Semantic representation, while increasingly adopting whole-text annotation rather than more specific
%shallow schemes, faces several challenges: semantic distinctions, besides requiring deeper understanding
%and being harder to learn, are not always well-defined, and progress is hindered by fragmentation.
%Semantic schemes diverge in the content they choose to annotate, their coupling with syntax,
%and their degree of cross-linguistic applicability and consistency \cite{abend2017state}.

Indeed, recent work in semantic parsing has targeted, among others,
Abstract Meaning Representation \cite[AMR;][]{banarescu2013abstract},
bilexical Semantic Dependencies \cite[SDP;][]{oepen2016towards}
%with target representations such as DELPH-IN MRS \cite[DM;][]{flickinger2012deepbank},
and Universal Conceptual Cognitive Annotation \cite[UCCA;][]{abend2013universal}.
While these schemes are formally different and focus on different distinctions,
much of their semantic content is shared \cite{abend2017state}.

Multitask learning \cite[MTL; ][]{caruana1998multitask} allows exploiting the overlap between tasks
to effectively extend the training data, 
and has greatly advanced with neural networks and representation learning
(see \S\ref{sec:related_work}).
We build on these ideas and propose a general transition-based DAG parser,
able to parse UCCA, AMR, SDP and UD \cite{nivre2016universal}.
We train the parser using MTL to obtain significant improvements
on UCCA parsing over single-task training in
(1) in-domain and (2) out-of-domain settings in English;
(3) an in-domain setting in German; and
(4) an in-domain setting in French, where training data is
scarce.

The novelty of this work is in proposing a general parsing and learning
architecture, able to accommodate such widely different parsing tasks, and in leveraging it
to show benefits from learning them jointly.

%%%%%%%%%%%%%%%%%%%%%%%%%%%%%%%%%%%%%%%%%%%%%%%%%%%%%%%%%%%%%%%%%%%%%%%%%%%%%%%%%
\section{Related Work}\label{sec:related_work}

MTL has been used over the years for NLP tasks with varying degrees of similarity,
examples including joint classification of different arguments in 
semantic role labeling \cite{toutanova2005joint},
and joint parsing and named entity recognition \cite{Finkel2009JointPA}.
%and  of unlabeled data for various tasks \cite{ando2005framework}.
Similar ideas, of parameter sharing across models trained with different datasets,
can be found in studies of domain adaptation \cite{W06-1615,P07-1033,K17-1040}.
For parsing, domain adaptation has been applied successfully in
parser combination and co-training \cite{mcclosky2010automatic,baucom2013domain}.

Neural MTL has mostly been effective in tackling formally similar
tasks \cite{P16-2038},
%or where one task tends to be used as pre-processing for another.
including
multilingual syntactic dependency parsing \cite{Q16-1031,guo2016exploiting},
as well as multilingual \cite{duong2017multilingual},
and cross-domain semantic parsing \cite{herzig-berant:2017:Short,W17-2607}.

Sharing parameters with a low-level task
has shown great benefit for transition-based syntactic parsing,
when jointly training with POS tagging
\cite{bohnet2012transition,Zhang2016StackpropagationIR}, and
with lexical analysis \cite{constant-nivre:2016:P16-1,more2016joint}.
Recent work has achieved state-of-the-art results in multiple NLP tasks
by jointly learning the tasks forming the NLP standard pipeline using 
a single neural model \cite{collobert2011natural,D17-1206},
thereby avoiding cascading errors, common in pipelines.

Much effort has been devoted to joint learning of syntactic
and semantic parsing, including
two CoNLL shared tasks \cite{surdeanu2008conll,hajivc2009conll}.
%on joint syntactic parsing and semantic role labeling.
Despite their conceptual and practical appeal, such joint models rarely outperform
the pipeline approach %of basing semantic parsing on the output of syntactic parsers
\cite{lluis2008joint,henderson2013multilingual,D15-1169,swayamdipta-EtAl:2016:CoNLL,swayamdipta2017frame}.

\citet{P17-1186} performed MTL for SDP in a closely related setting to ours.
They tackled three tasks, annotated over the same text
and sharing the same formal structures (bilexical DAGs),
with considerable edge overlap,
but differing in target representations (see \S\ref{sec:tasks}).
For all tasks, they reported an increase of 0.5-1 labeled $F_1$ points.
Recently, \citet{Peng-EtAl:2018:NAACL} applied a similar approach to
joint frame-semantic parsing and semantic dependency parsing,
using disjoint datasets, and reported further improvements.

%In this work, we take the challenge of MTL for semantic parsing a step further,
%improving UCCA parsing using MTL with conceptually and formally different tasks (AMR, SDP and UD; see \S\ref{sec:tasks}), 
%whose training sets are from different domains.
%Further, we show additional benefit from using a syntactic task parsing as auxiliary (UD).

\begin{figure}[!ht]
\fbox{\begin{subfigure}{0.47\textwidth}
  \centering
  \scalebox{.95}{
  \begin{tikzpicture}[level distance=8mm, ->]
    \node (ROOT) [fill=black, circle] {}
      child {node (After) {After} edge from parent node[above] {\scriptsize $L$}}
      child {node (graduation) [fill=black, circle] {}
      {
        child {node {graduation} edge from parent node[left] {\scriptsize $P$}}
      } edge from parent node[left] {\scriptsize $H$} }
      child {node {,} edge from parent node[below] {\scriptsize $U$}}
      child {node (moved) [fill=black, circle] {}
      {
        child {node (John) {John} edge from parent node[left] {\scriptsize $A$}}
        child {node {moved} edge from parent node[left] {\scriptsize $P$}}
        child {node [fill=black, circle] {}
        {
          child {node {to} edge from parent node[left] {\scriptsize $R$}}
          child {node {Paris} edge from parent node[right] {\scriptsize $C$}}
        } edge from parent node[above] {\scriptsize $A$} }
      } edge from parent node[above] {\scriptsize $H$} }
      ;
    \draw[dashed,->] (graduation) to node [above] {\scriptsize $A$} (John);
    \node (LKG) at (-1.8,0) [fill=cyan, circle] {};
    \draw[bend right,color=cyan] (LKG) to node [auto, left] {\scriptsize $LR$} (After);
    \draw[color=cyan] (LKG) to[out=-60, in=190] node [below] {\scriptsize $LA\quad$} (graduation);
    \draw[color=cyan] (LKG) to[out=30, in=90] node [above] {\scriptsize $LA$} (moved);
  \end{tikzpicture}
  }\caption{UCCA \label{fig:original_example_ucca}}
\end{subfigure}}

\fbox{\begin{subfigure}{0.47\textwidth}
  \centering
  \begin{tikzpicture}[->,
      every node/.append style={sloped,anchor=south,auto=false,font=\tiny},
      level 1/.style={level distance=14mm,sibling distance=26mm},
      level 2/.style={level distance=13mm},
      level 3/.style={level distance=12mm}]
    \node (ROOT) [draw=black,ellipse] {move-01}
      child {node [draw=black,ellipse] {after}
      {
            child {node (graduation) [draw=black,ellipse] {graduate-01} edge from parent node {op1} }
      } edge from parent node {time} }
      child {node (John) [draw=black,ellipse] {person}
      {
        child {node [draw=black,ellipse] {name}
        {
            child {node [draw=black,ellipse] {"John"} edge from parent node {op1} }
        } edge from parent node {name} }
      } edge from parent node {ARG0} }
      child {node [draw=black,ellipse] {city}
      {
        child {node [draw=black,ellipse] {name}
        {
            child {node [draw=black,ellipse] {"Paris"} edge from parent node {op1} }
        } edge from parent node {name} }
      } edge from parent node {ARG2} }
      ;
      \draw (graduation) to node {ARG0} (John);
  \end{tikzpicture}
  \caption{AMR \label{fig:original_example_amr}}
\end{subfigure}}

\fbox{\begin{subfigure}{0.47\textwidth}
  \centering
    \begin{dependency}[text only label, label style={above}, font=\small]
    \begin{deptext}[column sep=.8em,ampersand replacement=\^]
    After \^ graduation \^ , \^ John \^ moved \^ to \^ Paris \\
    \end{deptext}
        \depedge[edge unit distance=1ex]{1}{2}{ARG2}
        \depedge[edge unit distance=1ex]{5}{4}{ARG1}
        \depedge[edge unit distance=1ex, edge end x offset=-2pt]{1}{5}{ARG1}
        \deproot[edge unit distance=1.5ex]{5}{top}
        \depedge[edge unit distance=2ex, edge start x offset=-1pt, edge end x offset=3pt]{5}{7}{ARG2}
        \depedge[edge unit distance=1ex, edge end x offset=5pt]{6}{5}{ARG1}
        \depedge[edge unit distance=1ex]{6}{7}{ARG2}
    \end{dependency}
  \caption{DM \label{fig:original_example_sdp}}
\end{subfigure}}

\fbox{\begin{subfigure}{0.47\textwidth}
  \centering
    \begin{dependency}[text only label, label style={above}, font=\small]
    \begin{deptext}[column sep=.8em,ampersand replacement=\^]
    After \^ graduation \^ , \^ John \^ moved \^ to \^ Paris \\
    \end{deptext}
        \depedge[edge unit distance=1ex]{2}{1}{case}
        \depedge[edge unit distance=1ex]{4}{3}{punct}
        \depedge[edge unit distance=1ex]{5}{4}{nsubj}
        \depedge[edge unit distance=1ex, edge end x offset=-2pt]{2}{5}{obl}
        \depedge[edge unit distance=1ex]{7}{6}{case}
        \deproot[edge unit distance=1.5ex]{5}{root}
        \depedge[edge unit distance=1.5ex]{5}{7}{obl}
    \end{dependency}
  \caption{UD \label{fig:original_example_ud}}
\end{subfigure}}

\caption{\label{fig:original_examples}
 Example graph for each task.
 Figure \ref{fig:original_example_ucca} presents a UCCA graph. The dashed edge is remote,
  while the blue node and its outgoing edges represent inter-Scene linkage.
  Pre-terminal nodes and edges are omitted for brevity. 
 Figure \ref{fig:original_example_amr} presents an AMR graph.
  Text tokens are not part of the graph, and must be matched to
  concepts and constants by alignment.
  Variables are represented by their concepts.
 Figure \ref{fig:original_example_sdp} presents a DM semantic dependency graph,
  containing multiple roots: ``After'', ``moved'' and ``to'',
  of which ``moved'' is marked as \textit{top}.
  Punctuation tokens are excluded from SDP graphs.
 Figure \ref{fig:original_example_ud} presents a UD tree.
  %Each word has exactly one head, and there is a single root.
  Edge labels express syntactic relations.}

\end{figure}
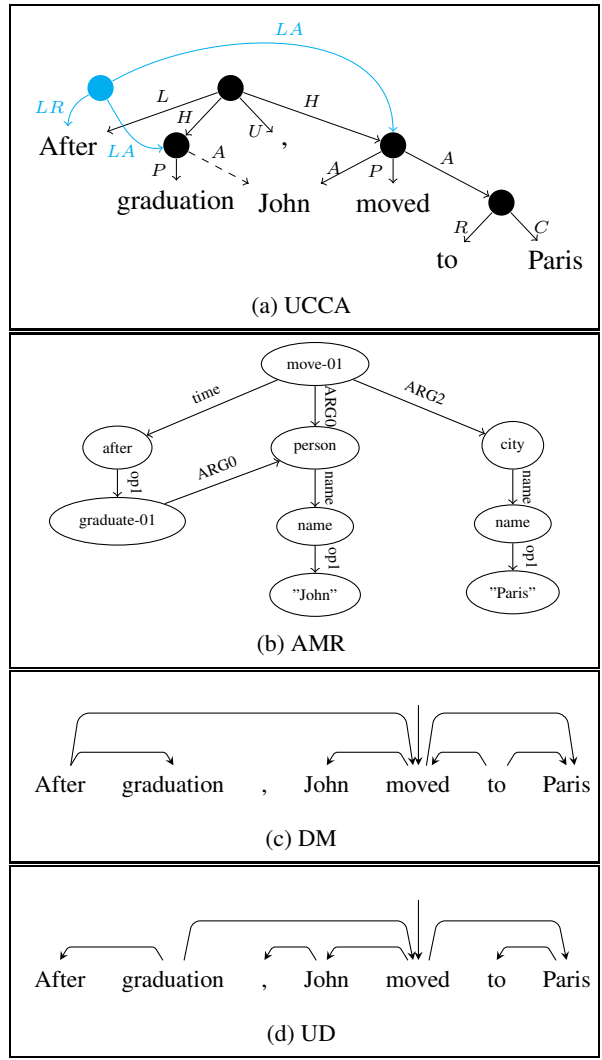

%%%%%%%%%%%%%%%%%%%%%%%%%%%%%%%%%%%%%%%%%%%%%%%%%%%%%%%%%%%%%%%%%%%%%%%%%%%%%%%%%
\section{Tackled Parsing Tasks}\label{sec:tasks}

In this section, we outline the parsing tasks we address.
We focus on representations that produce full-sentence analyses,
i.e., produce a graph covering all (content) words in the text, 
or the lexical concepts they evoke.
This contrasts with ``shallow'' semantic parsing,
primarily semantic role labeling
\cite[SRL;][]{gildea2002automatic,Palmer:05},
which targets argument structure phenomena using flat structures.
We consider four formalisms: UCCA, AMR, SDP and Universal Dependencies.
Figure~\ref{fig:original_examples} presents one sentence annotated in each scheme.
%annotating the sentence ``After graduation, John moved to Paris''.

%Although each of these tasks uses different graph structures,
%they all involve whole-sentence (or whole-paragraph) semantic analysis,
%and due to the structure of phenomena such as predicate-argument relations,
%require parsers that can handle non-projectivity (or discontinuity) and reentrancy, resulting in
%directed acyclic graphs (DAGs).\oa{maybe add a diagram or at least elaborate more on these like we
%did in the TUPA paper, with examples etc.}

\paragraph{Universal Conceptual Cognitive Annotation.}\label{sec:ucca}
UCCA \cite{abend2013universal} is a semantic representation whose main design principles
are ease of annotation, cross-linguistic applicability, and a modular architecture.
%of semantic distinctions.
UCCA represents the semantics of linguistic utterances
as directed acyclic graphs (DAGs), where terminal (childless) nodes
correspond to the text tokens, and non-terminal nodes to semantic units that participate
in some super-ordinate relation.
Edges are labeled, indicating the role of a child in the relation the parent represents.
Nodes and edges belong to one of several \textit{layers}, each corresponding
to a ``module'' of semantic distinctions.
UCCA's \textit{foundational layer} (the only layer for which annotated data exists)
mostly covers predicate-argument structure, semantic heads and inter-Scene relations.
%The \textit{linkage} layer covers relations between events, including temporal and discourse relations
%(exemplified by the gray node and its outgoing edges in Figure~\ref{fig:original_example_ucca}).

UCCA distinguishes \textit{primary} edges, corresponding 
to explicit relations, from \textit{remote} edges (appear dashed in
Figure~\ref{fig:original_example_ucca}) that allow for a unit to participate
in several super-ordinate relations.
Primary edges form a tree in each layer, whereas remote edges enable reentrancy, forming a DAG.
%As UCCA annotated data is currently fairly scarce (see \S\ref{sec:experiments}), 
%we hypothesize it will benefit from MTL, and consider it as our
%main task.

%%%%%%%%%%%%%%%%%%%%%%%%%%%%%%%%%%%%%%%%%%%%%%%%%%%%%%%%%%%%%%%%%%%%%%%%%%%%%5
\paragraph{Abstract Meaning Representation.}\label{sec:amr}

AMR \cite{banarescu2013abstract} is a semantic representation that encodes information about named entities, 
argument structure, semantic roles, word sense and co-reference.
AMRs are rooted directed graphs, in which both nodes and edges are labeled.
Most AMRs are DAGs, although cycles are permitted.

AMR differs from the other schemes we consider in that it does not anchor its graphs
in the words of the sentence (Figure~\ref{fig:original_example_amr}). Instead, AMR graphs
connect variables, concepts (from a pre-defined set) and constants (which may be strings or numbers).
Still, most AMR nodes are alignable to text tokens, a tendency used by AMR parsers,
which align a subset of the graph nodes to a subset of the text tokens (concept identification). In this work, we use pre-aligned AMR graphs.

Despite the brief period since its inception, AMR has been targeted by a number of works,
notably in two SemEval shared tasks \cite{may2016semeval,may2017semeval}.
To tackle its variety of distinctions and unrestricted graph structure,
AMR parsers often use specialized methods.
Graph-based parsers construct AMRs
by identifying concepts and scoring edges between them, either in a pipeline fashion
\cite{flanigan2014discriminative,artzi2015broad,pust2015parsing,foland2017abstract},
or jointly \cite{zhou2016amr}.
Another line of work %uses sequence-to-sequence models \cite{sutskever2014sequence},
trains machine translation models to convert strings into linearized AMRs
%treating AMR parsing as a machine translation model and linearizing the graph structure
\cite{barzdins2016riga,Gildea2017AddressingTD,Konstas2017NeuralAS,Buys2017RobustIN}.
Transition-based AMR parsers either 
use dependency trees as pre-processing, then mapping them into AMRs
\cite{wang-xue-pradhan:2015:ACL-IJCNLP,wang2015transition,wang-EtAl:2016:SemEval,goodman2016noise},
or use a transition system tailored to AMR parsing \cite{damonte-17,D17-1130}.
We differ from the above approaches in addressing AMR parsing 
using the same general DAG parser used for other schemes.

%%%%%%%%%%%%%%%%%%%%%%%%%%%%%%%%%%%%%%%%%%%%%%%%%%%%%%%%%%%%%%%%%%%%%%%%%%%%%5
\paragraph{Semantic Dependency Parsing.}\label{sec:sdp}

SDP uses a set of related representations, targeted in two recent SemEval shared tasks 
\cite{oepen2014semeval,oepen2015semeval}, and extended by \citet{oepen2016towards}.
They correspond to four semantic representation schemes, referred to as
DM, PAS, PSD and CCD, representing
predicate-argument relations between content words in a sentence.
All are based on semantic formalisms %whose annotation has been
converted into bilexical dependencies---directed graphs whose nodes are text tokens.
Edges are labeled, encoding semantic relations between the tokens.
Non-content tokens, such as punctuation,
are left out of the analysis (see Figure~\ref{fig:original_example_sdp}).
%but the sub-graph restricted to content-bearing tokens is connected.
Graphs containing cycles have been removed from the SDP datasets.

We use one of the representations
from the SemEval shared tasks: DM (DELPH-IN MRS), converted from 
DeepBank \cite{flickinger2012deepbank}, a corpus of hand-corrected parses from LinGO
ERG \cite{copestake2000open},
an HPSG \cite{pollard1994head}
using Minimal Recursion Semantics \cite{copestake2005minimal}.

%%%%%%%%%%%%%%%%%%%%%%%%%%%%%%%%%%%%%%%%%%%%%%%%%%%%%%%%%%%%%%%%
\paragraph{Universal Dependencies.}\label{sec:ud}
UD \cite{nivre2016universal,11234/1-2515} has quickly become
the dominant dependency scheme for
syntactic  annotation in many languages,
aiming for cross-linguistically consistent and coarse-grained treebank
annotation. Formally, UD uses bilexical trees, with edge labels 
representing syntactic relations between words.

We use UD as an auxiliary task,
inspired by previous work on joint syntactic and semantic parsing
(see \S\ref{sec:related_work}).
%\cite{lluis2008joint,collobert2011natural,D15-1169,swayamdipta-EtAl:2016:CoNLL,swayamdipta2017frame}.
In order to reach comparable analyses cross-linguistically,
UD often ends up in annotation that is similar to the common practice
in semantic treebanks, such as linking content words to content words wherever possible.
Using UD further allows conducting experiments on languages other than English, 
for which AMR and SDP annotated data is not available (\S\ref{sec:experiments}).

In addition to basic UD trees, we use the \textit{enhanced++} UD graphs available for English,
which are generated by the Stanford CoreNLP converters
\cite{SCHUSTER16.779}.\footnote{\url{http://github.com/stanfordnlp/CoreNLP}}
These include additional and augmented relations between content words,
partially overlapping with the notion of remote edges in UCCA:
in the case of control verbs, for example, a direct relation is added in 
enhanced++ UD between the subordinated verb and its controller,
which is similar to the semantic schemes' treatment of this construction.

%%%%%%%%%%%%%%%%%%%%%%%%%%%%%%%%%%%%%%%%%%%%%%%%%%%%%%%%%%%%%%%%%%%%%%%%%%%%%%%%%%%%%%%%
\section{General Transition-based DAG Parser}\label{sec:model}

All schemes considered in this work exhibit
reentrancy and discontinuity (or non-projectivity), to varying degrees.
In addition, UCCA and AMR contain non-terminal nodes.
To parse these graphs,
we extend TUPA \cite{hershcovich2017a}, 
a transition-based parser 
originally developed for UCCA,  as it
%is the only parser to our knowledge that
supports all these structural properties.
TUPA's transition system can yield any labeled DAG
whose terminals are anchored in the text tokens.
To support parsing into AMR, which uses graphs that are not anchored in the tokens,
 we take advantage of existing alignments of the graphs with the text
tokens during training (\S\ref{sec:format}).

First used for projective syntactic dependency tree parsing \cite{Nivre03anefficient},
transition-based parsers have since been generalized to parse into many other
graph families, such as (discontinuous) constituency trees \cite[e.g., ][]{zhang2009transition,maier-lichte:2016:DiscoNLP},
and DAGs \cite[e.g.,][]{sagae2008shift,du-EtAl:2015:SemEval}. %ribeyre-villemontedelaclergerie-seddah:2014:SemEval,
Transition-based parsers apply \textit{transitions}
incrementally to an internal state defined by a buffer $B$ of remaining tokens 
and nodes, a stack $S$ of unresolved nodes, and a labeled graph $G$ of 
constructed nodes and edges.
When a terminal state is reached, the graph $G$ is the final output.
A classifier is used at each step to select the next transition, 
based on features that encode the current state.
%During training, an oracle converts the gold-standard annotations into
% training instances.

\subsection{TUPA's Transition Set}\label{sec:transition_set}

Given a sequence of tokens $w_1, \ldots, w_n$,
we predict a rooted graph $G$ whose terminals are the tokens.
Parsing starts with the root node on the stack,
and the input tokens in the buffer.

The TUPA transition set includes
the standard \textsc{Shift} and \textsc{Reduce} operations,
\textsc{Node$_X$} for creating a new non-terminal node and an $X$-labeled edge,
\textsc{Left-Edge$_X$} and \textsc{Right-Edge$_X$} to create a new primary $X$-labeled edge,
\textsc{Left-Remote$_X$} and \textsc{Right-Remote$_X$} to create a new remote $X$-labeled edge,
\textsc{Swap} to handle discontinuous nodes,
and \textsc{Finish} to mark the state as terminal.

Although UCCA contains nodes without any text tokens as descendants
(called \textit{implicit units}),
these nodes are infrequent and only cover 0.5\% of non-terminal nodes.
For this reason we follow previous work \cite{hershcovich2017a} and discard implicit units from
the training and evaluation,
and so do not include transitions for creating them.

In AMR, implicit units are considerably more common, as any unaligned concept
with no aligned descendents is implicit (about 6\% of the nodes).
Implicit AMR nodes usually result from alignment errors, or from abstract concepts
which have no explicit realization in the text \cite{buys2017oxford}.
We ignore implicit nodes when training on AMR as well.
TUPA also does not support node labels, 
which are ubiquitous in AMR but absent in UCCA structures (only edges are labeled in UCCA). 
We therefore only produce edge labels and not node labels when training on AMR.

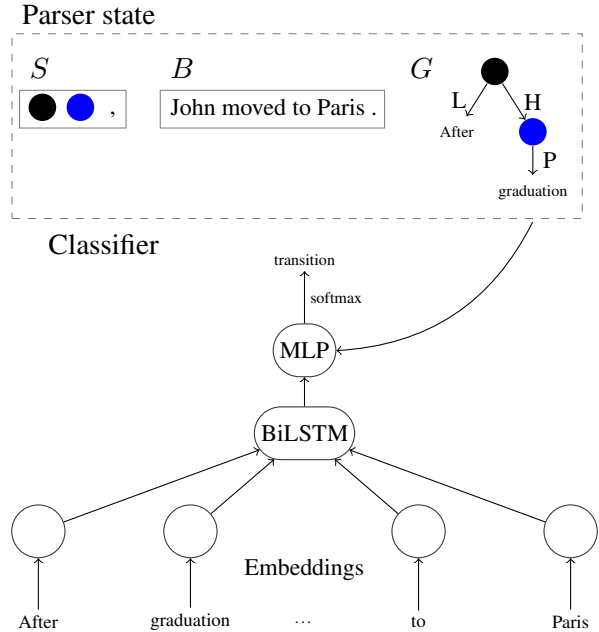
\begin{figure}[t]
   \begin{tikzpicture}[level distance=8mm, sibling distance=1cm]
   \node[anchor=west] at (0,1.5) {Parser state};
   \draw[color=gray,dashed] (0,-1.2) rectangle (7.5,1.25);
   \draw[color=gray] (.1,0) rectangle (1.5,.5);
   \node[anchor=west] at (.1,.8) {$S$};
   \node[fill=black, circle] at (.4,.275) {};
   \node[fill=blue, circle] at (.9,.275) {};
   \node[anchor=west] at (1.15,.175) {\small ,};
   \draw[color=gray] (1.95,0) rectangle (4.9,.5);
   \node[anchor=west] at (1.95,.8) {$B$};
   \node[anchor=west] at (1.95,.275) {\small John moved to Paris .};
   \node[anchor=west] at (5.1,.8) {$G$};
   \node[fill=black, circle] at (6.35,.75) {}
     child {node  {\tiny After} edge from parent [->] node[left] {\small L}}
     child {node [fill=blue, circle] {}
     {
       child {node {\tiny graduation} edge from parent [->] node[right] {\small P}}
     } edge from parent [->] node[right] {\small H} };
   \end{tikzpicture}
   \begin{tikzpicture}[->]
   \node[anchor=west] at (0,6) {Classifier};
   \tiny
   \tikzstyle{main}=[rounded rectangle, minimum size=7mm, draw=black!80, node distance=12mm]
   \node[main] (specific) at (3.5,3.5) {\small BiLSTM};
   \node (embeddings) at (3.5,1.7) {\small Embeddings};
   \foreach \i/\word in {0/{After},2/{graduation},5/{to},7/{Paris}} {
       \node (x\i) at (\i,1) {\scriptsize \word};
       \node[main] (e\i) at (\i,2.2) {};
       \path (x\i) edge (e\i);
       \path (e\i) edge (specific);
   }
    \node (x4) at (3.5,1) {\ldots};
    \node[main] (mlp) at (3.5,4.6) {\small MLP};
    \path (specific) edge (mlp);
    \coordinate (state) at (6.5,6.3);
    \path (state) edge [bend left] (mlp);
    \node (transition) at (3.5,5.8) {transition};
    \path (mlp) edge node[right] {softmax} (transition);
   \end{tikzpicture}
\caption{Illustration of the TUPA model, adapted from \citet{hershcovich2017a}.
Top: parser state.
Bottom: BiLTSM architecture.}\label{fig:single_model}
\end{figure}

%%%%%%%%%%%%%%%%%%%%%%%%%%%%%%%%%%%%%%%%%%%%%%%%%%%%%%%%%%%%%%%%%%%%%%%%%%%%%%%%%
\subsection{Transition Classifier}\label{sec:classifier}

To predict the next transition at each step,
we use a BiLSTM with embeddings as inputs,
followed by an MLP and a softmax layer for classification \cite{kiperwasser2016simple}.
The model is illustrated in Figure~\ref{fig:single_model}.
Inference is performed greedily,
and training is done with an oracle that yields the set of all optimal 
transitions at a given state (those that lead to a state from which the gold graph is still reachable).
Out of this set, the actual transition performed in training is the one
with the highest score given by the classifier,
which is trained to maximize the sum of log-likelihoods of all 
optimal transitions at each step.

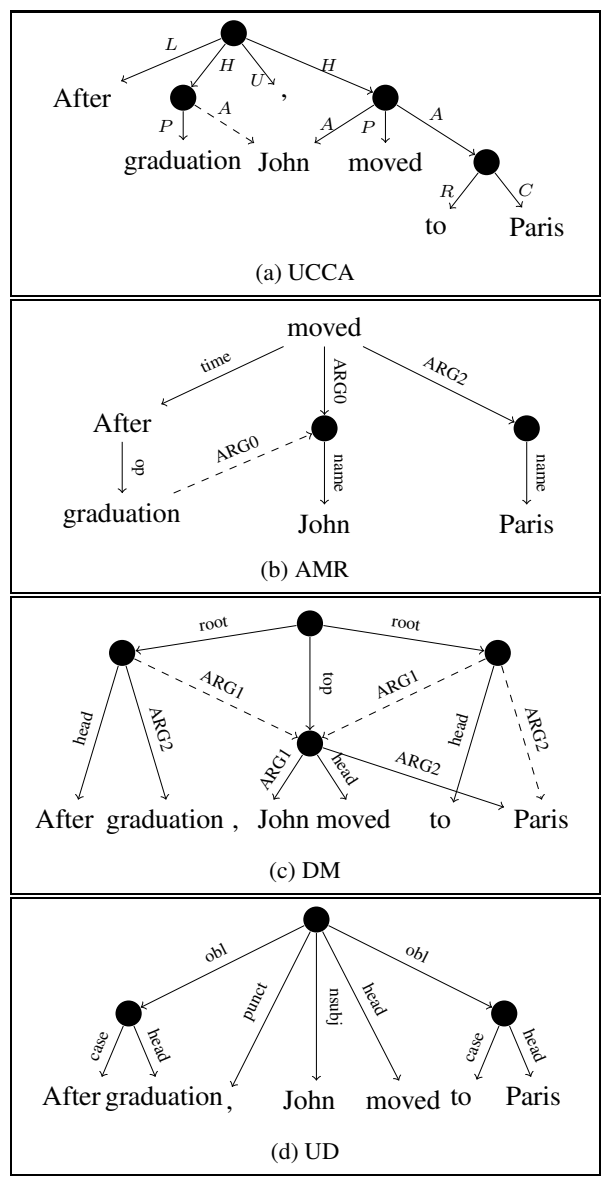
\begin{figure}[!ht]
\fbox{\begin{subfigure}{0.47\textwidth}
    \centering
    \scalebox{.95}{
    \begin{tikzpicture}[level distance=9mm, sibling distance=14mm, ->,
        every circle node/.append style={fill=black}]
      \tikzstyle{word} = [font=\rmfamily,color=black]
      \node (ROOT) [circle] {}
        child {node (After) [word] {After} edge from parent node[above] {\scriptsize $L$}}
        child {node (graduation) [circle] {}
        {
          child {node [word] {graduation} edge from parent node[left] {\scriptsize $P$}}
        } edge from parent node[right] {\scriptsize $H$} }
        child {node [word] {,} edge from parent node[below] {\scriptsize $U$}}
        child {node (moved) [circle] {}
        {
          child {node (John) [word] {John} edge from parent node[left] {\scriptsize $A$}}
          child {node [word] {moved} edge from parent node[left] {\scriptsize $P$}}
          child {node [circle] {}
          {
            child {node [word] {to} edge from parent node[left] {\scriptsize $R$}}
            child {node [word] {Paris} edge from parent node[right] {\scriptsize $C$}}
          } edge from parent node[above] {\scriptsize $A$} }
        } edge from parent node[right] {\scriptsize $H$} }
        ;
      \draw[dashed,->] (graduation) to node [above] {\scriptsize $A$} (John);
    \end{tikzpicture}}
  \caption{UCCA}
  \label{fig:converted_example_ucca}
\end{subfigure}}

\fbox{\begin{subfigure}{0.47\textwidth}
  \centering
  \scalebox{.95}{
  \begin{tikzpicture}[level distance=16mm, ->,
      every node/.append style={sloped,anchor=south,auto=false,font=\scriptsize},
      level 1/.style={sibling distance=28mm},
      level 2/.style={sibling distance=14mm},
      level 3/.style={sibling distance=12mm}]
    \tikzstyle{word} = [font=\rmfamily,color=black]
    \node (ROOT) [word] {moved}
      child {node [word] {After}
      {
            child {node (graduation) [word] {graduation} edge from parent node {op} }
      } edge from parent node {time} }
      child {node (John) [fill=black,circle] {}
      {
        child {node [word] {John} edge from parent node {name} }
      } edge from parent node {ARG0} }
      child {node [fill=black,circle] {}
      {
        child {node [word] {Paris} edge from parent node {name} }
      } edge from parent node {ARG2} }
      ;
      \draw[dashed] (graduation) to node {ARG0} (John);
  \end{tikzpicture}}
  \captionof{figure}{AMR}
  \label{fig:converted_example_amr}
\end{subfigure}}

\fbox{\begin{subfigure}{0.47\textwidth}
  \centering
  \scalebox{.95}{
  \begin{tikzpicture}[level distance=12mm, ->,
      every node/.append style={sloped,anchor=south,auto=false,font=\scriptsize},
      level 1/.style={sibling distance=26mm,level distance=6mm},
      level 2/.style={sibling distance=14mm,level distance=14mm}]
    \tikzstyle{word} = [font=\rmfamily,color=black]
    \node (ROOT) [fill=black,circle] {}
      child {node (after) [fill=black,circle] {}
      {
        child {node [draw=none] {}
        {
          child {node [word] (after_word) {After{\color{white}g}} edge from parent [draw=none]}
        } edge from parent [draw=none] }
        child {node [draw=none] {}
        {
          child {node [word] (graduation) {graduation ,} edge from parent [draw=none]}
        } edge from parent [draw=none] }
      } edge from parent node {root}}
      child {node [draw=none] {}
      {
        child {node (moved) [fill=black,circle] {}
        {
          child {node [word] {\quad{\color{white}g} John} edge from parent node {ARG1}}
          child {node [word] {moved{\color{white}g}} edge from parent node {head}}
        } edge from parent [draw=none] }
      } edge from parent [draw=none] }
      child {node (to) [fill=black,circle] {}
      {
        child {node [draw=none] {}
        {
            child {node [word] (to_word) {to{\color{white}g}} edge from parent [draw=none]}
          } edge from parent [draw=none] }
          child {node [draw=none] {}
        {
          child {node [word] (Paris) {Paris{\color{white}g}} edge from parent [draw=none]}
        } edge from parent [draw=none] }
      } edge from parent node {root}}
      ;
      \draw (ROOT) to node {top} (moved);
      \draw (after) to node {head} (after_word);
      \draw (after) to node {ARG2} (graduation);
      \draw[dashed] (after) to node {ARG1} (moved);
      \draw[dashed] (to) to node {ARG1} (moved);
      \draw (to) to node {head} (to_word);
      \draw (moved) to node {ARG2} (Paris);
      \draw[dashed] (to) to node {ARG2} (Paris);
  \end{tikzpicture}}
  \captionof{figure}{DM}
  \label{fig:converted_example_sdp}
\end{subfigure}}

\fbox{\begin{subfigure}{0.47\textwidth}
  \centering
  \scalebox{.95}{
  \begin{tikzpicture}[level distance=15mm, ->,
      every node/.append style={sloped,anchor=south,auto=false,font=\scriptsize},
      level 1/.style={sibling distance=13mm},
      level 2/.style={sibling distance=1cm}]
    \tikzstyle{word} = [font=\rmfamily,color=black]
    \node (ROOT) [fill=black,circle] {}
      child {node (after) [fill=black,circle] {}
      {
        child {node [word] {After{\color{white}g}\quad\quad} edge from parent node {case}}
        child {node [word] {\quad graduation\quad\quad} edge from parent node {head}}
      } edge from parent node {obl}}
      child {node {}
      {
        child {node [word] (comma) {\quad,{\color{white}g}} edge from parent [draw=none]}
      } edge from parent [draw=none]}
      child {node {}
      {
        child {node [word] (John) {John{\color{white}g}} edge from parent [draw=none]}
      } edge from parent [draw=none]}
      child {node {}
      {
        child {node [word] (moved) {moved{\color{white}g}} edge from parent [draw=none]}
      } edge from parent [draw=none]}
      child {node (to) [fill=black,circle] {}
      {
          child {node [word] {to{\color{white}g}} edge from parent node {case}}
          child {node [word] {Paris{\color{white}g}} edge from parent node {head}}
      } edge from parent node {obl}}
      ;
      \draw (ROOT) to node {punct} (comma);
      \draw (ROOT) to node {nsubj} (John);
      \draw (ROOT) to node {head} (moved);
  \end{tikzpicture}}
  \captionof{figure}{UD}\label{fig:converted_example_ud}
\end{subfigure}}

\caption{Graphs from Figure~\ref{fig:original_examples}, after conversion to  the unified DAG format
(with pre-terminals omitted: each terminal drawn in place of its parent).
Figure~\ref{fig:converted_example_ucca} presents a converted UCCA graph.
Linkage nodes and edges are removed, but the original graph is otherwise preserved.
Figure~\ref{fig:converted_example_amr} presents a converted AMR graph, with
text tokens added according to the alignments.
Numeric suffixes of \textit{op} relations are removed,
and names collapsed.
Figure~\ref{fig:converted_example_sdp} presents a converted SDP graph (in the DM representation), with
intermediate non-terminal \textit{head} nodes introduced.
In case of reentrancy, an arbitrary reentrant edge is marked as remote.
Figure~\ref{fig:converted_example_ud} presents a converted UD graph.
As in SDP, intermediate non-terminals and \textit{head} edges are introduced.
While converted UD graphs form trees, enhanced++ UD graphs may not.}\label{fig:converted_examples}
\end{figure}

\paragraph{Features.}
We use the original TUPA features,
representing the words, POS tags, syntactic dependency relations, and previously predicted edge labels
for nodes in specific locations in the parser state.
In addition, for each token
we use embeddings representing the one-character prefix, three-character suffix,
shape (capturing orthographic features, e.g., ``Xxxx''),
and named entity type,\footnote{See Supplementary Material for a full listing of features.}
all provided by spaCy \cite{spacy2}.\footnote{\url{http://spacy.io}}
To the learned word vectors, we concatenate the 250K most frequent word vectors from fastText
\cite{bojanowski2016enriching},\footnote{\url{http://fasttext.cc}}
pre-trained over Wikipedia and updated during training.
%For AMR we add node label features according to gold node labels.

\paragraph{Constraints.}
As each annotation scheme has different constraints on the allowed graph structures,
we apply these constraints separately for each task.
During training and parsing, the relevant constraint set rules out some of the transitions
according to the parser state.
Some constraints are task-specific, others are generic.
For example, in UCCA, a terminal may only have one parent.
In AMR, a concept corresponding to a PropBank frame may only have 
the core arguments defined for the frame as children.
An example of a generic constraint is that stack nodes 
that have been swapped
should not be swapped again.\footnote{
 To implement this constraint, we define a \textit{swap index}
 for each node, assigned when the node is created.
 At initialization, only the root node and terminals exist.
 We assign the root a swap index of 0, and for each terminal, its
 position in the text (starting at 1).
 Whenever a node is created as a result of a \textsc{Node}
 transition, its swap index is the arithmetic
 mean of the swap indices of the stack top and buffer head.}

%%%%%%%%%%%%%%%%%%%%%%%%%%%%%%%%%%%%%%%%%%%%%%%%%%%%%%%%%%%%%%%%%%%%%%%%%%%%%%%%%%%%%%%%
\section{Unified DAG Format}\label{sec:format}

To apply our parser to the four target tasks (\S\ref{sec:tasks}),
we convert them into a unified DAG format, which is inclusive enough to
allow representing any of the schemes with very little loss of information.\footnote{See
Supplementary Material for more conversion details.}

The format consists of a rooted DAG, where the tokens are the terminal nodes.
As in the UCCA format, edges are labeled (but not nodes),
and are divided into \textit{primary} and \textit{remote} edges,
where the primary edges form a tree (all nodes have at most one primary parent,
and the root has none).
Remote edges enable reentrancy, and thus together with primary edges
form a DAG.
Figure~\ref{fig:converted_examples} shows examples for converted graphs.
Converting UCCA into the unified format consists simply of removing linkage 
nodes and edges (see Figure~\ref{fig:converted_example_ucca}), which were
also discarded by \citet{hershcovich2017a}.

\paragraph{Converting bilexical dependencies.}
To convert DM and UD into the unified DAG format,
we add a pre-terminal for each token,
and attach the pre-terminals according to the original dependency edges:
traversing the tree from the root down, for each head token we create a non-terminal
parent with the edge label {\it head},
and add the node's dependents as children of the created non-terminal node
(see Figures~\ref{fig:converted_example_sdp} and \ref{fig:converted_example_ud}).
Since DM allows multiple roots, we form a single root node, whose children
are the original roots. The added edges are labeled \textit{root}, where
top nodes are labeled \textit{top} instead.
In case of reentrancy, an arbitrary parent is marked as primary, and the rest as remote
(denoted as dashed edges in Figure~\ref{fig:converted_examples}).

\paragraph{Converting AMR.}
In the conversion from AMR, node labels are dropped.
Since alignments are not part of the AMR graph (see Figure~\ref{fig:converted_example_amr}),
we use automatic alignments (see \S\ref{sec:experiments}),
and attach each node with an edge to each of its aligned terminals.

%Since the order of AMR ordinal relations, such as \textit{op1}, \textit{op2},
%is annotated according to the order of text tokens,
%the numeric index is redundant and is thus removed.
%Numeric suffixes are kept when they are meaningful, e.g. in distinguishing between PropBank semantic
%roles (\textit{ARG[0-5]}).
Named entities in AMR are represented as a subgraph, whose \textit{name}-labeled root
has a child for each token in the name (see the two \textit{name} nodes in Figure~\ref{fig:original_example_amr}).
We collapse this subgraph into a single node whose children are the name tokens.

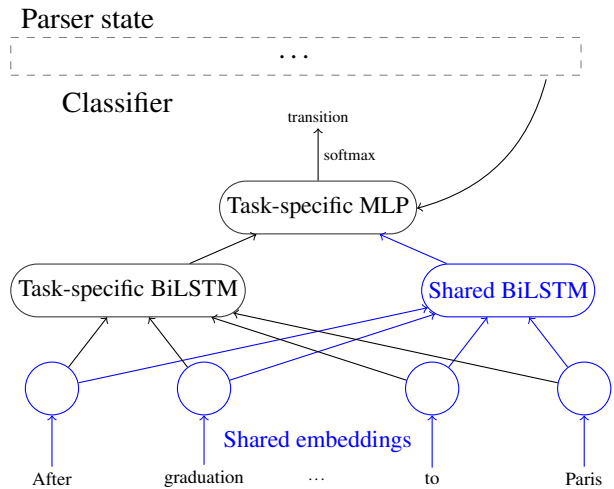
\begin{figure}[t]
   \begin{tikzpicture}
   \node[anchor=west] at (0,.75) {Parser state};
   \draw[color=gray,dashed] (0,0) rectangle (7.5,.5);
    \node (x4) at (3.75,0.25) {\ldots};
   \end{tikzpicture}
   \begin{tikzpicture}[->]
   \node[anchor=west] at (0,6) {Classifier};
   \tiny
   \tikzstyle{main}=[rounded rectangle, minimum size=7mm, draw=black!80, node distance=12mm]
   \node[main] (specific) at (1,3.5) {\small Task-specific BiLSTM};
   \node[main,color=blue] (shared) at (6,3.5) {\small Shared BiLSTM};
   \node[color=blue] (embeddings) at (3.5,1.5) {\small Shared embeddings};
   \foreach \i/\word in {0/{After},2/{graduation},5/{to},7/{Paris}} {
       \node (x\i) at (\i,1) {\scriptsize \word};
       \node[main,color=blue] (e\i) at (\i,2.2) {};
       \path[color=blue] (x\i) edge (e\i);
       \path (e\i) edge (specific);
       \path[color=blue] (e\i) edge (shared);
   }
    \node (x4) at (3.5,1) {\ldots};
    \node[main] (mlp) at (3.5,4.6) {\small Task-specific MLP};
    \path (specific) edge (mlp);
    \path[color=blue] (shared) edge (mlp);
    \coordinate (state) at (6.5,6.3);
    \path (state) edge [bend left] (mlp);
    \node (transition) at (3.5,5.8) {transition};
    \path (mlp) edge node[right] {softmax} (transition);
   \end{tikzpicture}
   \caption{MTL model.
      Token representations are computed both by a task-specific and a shared BiLSTM.
      Their outputs are concatenated with the parser state embedding,
      identical to Figure~\ref{fig:single_model},
      and fed into the task-specific MLP for selecting the next transition.
      Shared parameters are shown in blue.}\label{fig:multi_model}
\end{figure}

%%%%%%%%%%%%%%%%%%%%%%%%%%%%%%%%%%%%%%%%%%%%%%%%%%%%%%%%%%%%%%%%%%%%%%%%%%%%%%%%%%%%%%%%%%%%
\section{Multitask Transition-based Parsing}\label{sec:multitask}

Now that the same model can be applied to different tasks, 
we can train it in a multitask setting.
The fairly small training set available for UCCA (see \S\ref{sec:experiments})
makes MTL particularly appealing,
and we focus on it in this paper, treating AMR, DM and UD parsing as auxiliary tasks.

Following previous work, we share only some of the parameters
\cite{N16-1179,P16-2038,C16-1013,C16-1059,C16-1179,E17-1005,P17-1186,Peng-EtAl:2018:NAACL},
leaving task-specific
sub-networks as well.
Concretely, we keep the BiLSTM used by TUPA for the main task (UCCA parsing),
add a BiLSTM that is shared across all tasks,
and replicate the MLP (feedforward sub-network) for each task.
The BiLSTM outputs (concatenated for the main task) are
fed into the task-specific MLP (see Figure~\ref{fig:multi_model}).
Feature embeddings are shared across tasks.

\paragraph{Unlabeled parsing for auxiliary tasks.}
To simplify the auxiliary tasks and facilitate generalization \cite{E17-2026},
we perform unlabeled parsing for AMR, DM and UD,
while still predicting edge labels in UCCA parsing.
To support unlabeled parsing, we simply remove all labels from the
\textsc{Edge}, \textsc{Remote} and \textsc{Node} transitions output by the oracle.
This results in a much smaller number of transitions the classifier has to select from
(no more than 10, as opposed to 45 in labeled UCCA parsing),
allowing us to use no BiLSTMs and fewer dimensions and layers for task-specific MLPs
of auxiliary tasks (see \S\ref{sec:experiments}).
This limited capacity forces the network to use the shared parameters for all tasks,
increasing generalization \cite{E17-1005}.

\section{Experimental Setup}\label{sec:experiments}

We here detail a range of experiments to assess the value of MTL to UCCA parsing,
training the parser in single-task and multitask settings,
and evaluating its performance on the UCCA test sets in both in-domain and out-of-domain settings.

\begin{table*}[t]
\centering
\small
\setlength\tabcolsep{2pt}
\begin{tabular}{l|ccc|ccc||ccc|ccc||ccc|ccc}
& \multicolumn{6}{c||}{\footnotesize \bf English}
& \multicolumn{6}{c||}{\footnotesize \bf French}
& \multicolumn{6}{c}{\footnotesize \bf German} \\
\hline
& \multicolumn{3}{c|}{\footnotesize \bf {\#} tokens}
& \multicolumn{3}{c||}{\footnotesize \bf {\#} sentences}
& \multicolumn{3}{c|}{\footnotesize \bf {\#} tokens}
& \multicolumn{3}{c||}{\footnotesize \bf {\#} sentences}
& \multicolumn{3}{c|}{\footnotesize \bf {\#} tokens}
& \multicolumn{3}{c}{\footnotesize \bf {\#} sentences} \\
& \footnotesize \bf train & \footnotesize \bf dev & \footnotesize \bf test
& \footnotesize \bf train & \footnotesize \bf dev & \footnotesize \bf test
& \footnotesize \bf train & \footnotesize \bf dev & \footnotesize \bf test 
& \footnotesize \bf train & \footnotesize \bf dev & \footnotesize \bf test
& \footnotesize \bf train & \footnotesize \bf dev & \footnotesize \bf test
& \footnotesize \bf train & \footnotesize \bf dev & \footnotesize \bf test \\
\hline
\textbf{UCCA} &&&&&&&&&&&&&&&& \\
Wiki & 128444 & 14676 & 15313 & 4268 & 454 & 503 &&&&&&&&&&&& \\
20K &&& 12339 &&& 506 & 10047 & 1558 & 1324 & 413 & 67 & 67 & 79894 & 10059 & 42366 & 3429 & 561 & 2164 \\
\textbf{AMR} & \multicolumn{2}{l}{648950} && \multicolumn{2}{l}{36521} &&&&&&&&&&&&& \\
\textbf{DM} & \multicolumn{2}{l}{765025} && \multicolumn{2}{l}{33964} &&&&&&&&&&&&& \\
\textbf{UD} & \multicolumn{2}{l}{458277} && \multicolumn{2}{l}{17062} &&
\multicolumn{2}{l}{899163} && \multicolumn{2}{l}{32347} && \multicolumn{2}{l}{268145} && 13814
\end{tabular}
\caption{Number of tokens and sentences in the training, development and test sets
we use for each corpus and language.\label{tab:corpora}}
\end{table*}

\paragraph{Data.}

For UCCA, we use v1.2 of the English Wikipedia corpus \cite[\textit{Wiki};][]{abend2013universal},
with the standard train/dev/test split (see Table~\ref{tab:corpora}),
and the \textit{Twenty Thousand Leagues Under the Sea} corpora
\cite[\textit{20K};][]{sulem2015conceptual},
annotated in English, French and German.\footnote{\mbox{\url{http://github.com/huji-nlp/ucca-corpora}}}
For English and French we use 20K v1.0,
a small parallel corpus comprising the first five chapters of the book.
As in previous work \cite{hershcovich2017a}, we use the English part only as an out-of-domain test set.
We train and test on the French part using the standard split,
as well as the German corpus (v0.9),
which is a pre-release and still contains a considerable amount of noisy annotation.
Tuning is performed on the respective development sets.

For AMR, we use LDC2017T10, identical to the dataset targeted in SemEval 2017
\cite{may2017semeval}.\footnote{\mbox{\url{http://catalog.ldc.upenn.edu/LDC2017T10}}}
For SDP, we use the DM representation from the SDP 2016 dataset
\cite{oepen2016towards}.\footnote{\url{http://sdp.delph-in.net/osdp-12.tgz}}
For Universal Dependencies, we use all English, French and German treebanks from UD v2.1
\cite{11234/1-2515}.\footnote{\url{http://hdl.handle.net/11234/1-2515}}
We use the enhanced++ UD representation \cite{SCHUSTER16.779} in our English experiments,
henceforth referred to as UD$^{++}$.
We use only the AMR, DM and UD training sets from standard splits.

While UCCA is annotated over Wikipedia and over a literary corpus,
the domains for AMR, DM and UD are blogs, news, emails, reviews, and Q\&A.
This domain difference between training and test is particularly challenging (see \S\ref{sec:discussion}).
Unfortunately, none of the other schemes have available annotation over Wikipedia text.

\paragraph{Settings.}

We explore the following settings: 
(1) in-domain setting in English, training and testing on Wiki;
(2) out-of-domain setting in English, training on Wiki and testing on 20K;
(3) French in-domain setting, where available training dataset is small,
training and testing on 20K;
(4) German in-domain setting on 20K, with somewhat noisy annotation.
For MTL experiments, we use unlabeled AMR, DM and UD$^{++}$ parsing as auxiliary tasks in English,
and unlabeled UD parsing in French and German.\footnote{We did not use AMR, DM or UD$^{++}$ in French
and German, as these are only available in English.}
We also report baseline results training only the UCCA training sets.

\paragraph{Training.}

We create a unified corpus for each setting, shuffling all sentences from relevant datasets together,
but using only the UCCA development set $F_1$ score as the early stopping criterion.
In each training epoch, we use the same number of examples from each task---the
UCCA training set size.
Since training sets differ in size, we sample this many sentences from each one.
The model is implemented using DyNet \cite{neubig2017dynet}.\footnote{\url{http://dynet.io}}

\begin{table}[h]
\centering
\small
\setlength\tabcolsep{2pt}
\begin{tabular}{l|c|ccccc}
&& \multicolumn{3}{c}{\bf Multitask} \\ 
\bf Hyperparameter &  \bf Single & \bf Main & \bf Aux & \bf Shared \\
\hline
Pre-trained word dim. & 300 &&& 300 \\
Learned word dim. & 200 &&& 200 \\
POS tag dim. & 20 &&& 20 \\
Dependency relation dim. & 10 &&& 10 \\
Named entity dim. & 3 &&& 3 \\
Punctuation dim. & 1 &&& 1 \\
Action dim. & 3 &&& 3 \\
Edge label dim. & 20 & 20 \\
\hline
MLP layers & 2 & 2 & 1 \\
MLP dimensions & 50 & 50 & 50 \\
BiLSTM layers & 2 & 2 & & 2 \\
BiLSTM dimensions & 500 & 300 & & 300
\end{tabular}
\caption{Hyperparameter settings.
Middle column shows hyperparameters used for the single-task architecture,
described in \S\ref{sec:classifier}, and
right column for the multitask architecture,
described in \S\ref{sec:multitask}.
\textbf{Main} refers to parameters specific to the main task---UCCA parsing
(task-specific MLP and BiLSTM, and edge label embedding),
\textbf{Aux} to parameters specific to each auxiliary task
(task-specific MLP, but no edge label embedding since the tasks are unlabeled),
and \textbf{Shared} to parameters shared among all tasks
(shared BiLSTM and embeddings).\label{tab:hyperparams}}
\end{table}

\paragraph{Hyperparameters.}

We initialize embeddings randomly.
We use dropout \cite{srivastava2014dropout} between MLP layers, and recurrent dropout
\cite{NIPS2016_6241} between BiLSTM layers, both with $p=0.4$.
We also use word ($\alpha=0.2$), tag ($\alpha=0.2$) and dependency relation ($\alpha=0.5$) dropout
\cite{kiperwasser2016simple}.\footnote{In training, the embedding for a feature value $w$ is
replaced with a zero vector with a probability of $\frac{\alpha}{\#(w)+\alpha}$,
where $\#(w)$ is the number of occurrences of $w$ observed.}
In addition, we use a novel form of dropout, \textit{node dropout}:
with a probability of 0.1 at each step, all features associated with a single
node in the parser state are replaced with zero vectors.
For optimization we use a minibatch size of 100, decaying all weights by $10^{-5}$ at each update,
and train with stochastic gradient descent for $N$ epochs with a learning
rate of 0.1, followed by AMSGrad \cite{j.2018on} for $N$ epochs with
$\alpha=0.001,\beta_1=0.9$ and $\beta_2=0.999$.
We use $N=50$ for English and German, and $N=400$ for French.
We found this training strategy better than using only one of the optimization methods,
similar to findings by \citet{keskar2017improving}.
We select the epoch with the best average labeled $F_1$ score on the
UCCA development set.
Other hyperparameter settings are listed in Table~\ref{tab:hyperparams}.

\paragraph{Evaluation.}

We evaluate on UCCA using labeled precision, recall and $F_1$ 
on primary and remote edges,
following previous work \cite{hershcovich2017a}.
Edges in predicted and gold graphs are matched by terminal yield and label.
Significance testing of improvements over the single-task model is done
by the bootstrap test \cite{berg2012empirical}, with $p<0.05$.

%\subsection{Ensembling}
%
%During inference, we use Product of Experts \cite[PoE; ][]{hinton2002training} to combine the predictions
%of three models trained in the same setting, but with different random seeds. The transition selected is
%\[
%\argmax_{t\in T}\sum_{i=1}^3\big[\log(\mathrm{softmax}(m_i(s)))\big]_t
%\]
%where $T$ is the set of possible transitions, $m_i$ are the combined models, and $s$ is the current state.
%
%In order to ensemble multitask models, we combine models trained with the same auxiliary task.
%Another alternative is to combine models trained with different auxiliary tasks.\oa{don't write about alternatives,
%but rather on what we actually do}
%This provides greater variability between the combined models.

\section{Results}\label{sec:results}

\begin{table}[t]
\centering
\small
\setlength\tabcolsep{3pt}
\begin{tabular}{l|lll|lll}
& \multicolumn{3}{c|}{\footnotesize \bf Primary} & \multicolumn{3}{c}{\footnotesize \bf Remote} \\
& \footnotesize \textbf{LP} & \footnotesize \textbf{LR} & \footnotesize \textbf{LF}
& \footnotesize \textbf{LP} & \footnotesize \textbf{LR} & \footnotesize \textbf{LF} \\
\hline
\multicolumn{4}{l|}{\small \bf English (in-domain)} & \\
\footnotesize HAR17
& 74.4 & 72.7 & 73.5 & 47.4 & 51.6 & 49.4 \\
\footnotesize Single
& 74.4 & 72.9 & 73.6 & 53 & 50 & 51.5 \\
\cline{1-1}
%\small \bf Multitask &&& \\
\footnotesize AMR
& 74.7 & 72.8 & 73.7 & 48.7$\star$ & 51.1 & 49.9 \\
\footnotesize DM
& 75.7$\star$ & 73.9$\star$ & 74.8$\star$ & 54.9 & 53 & \textbf{53.9} \\
\footnotesize UD$^{++}$
& 75$\star$ & 73.2 & 74.1$\star$ & 49 & 52.7 & 50.8 \\
\footnotesize AMR + DM
& 75.6$\star$ & 73.9$\star$ & 74.7$\star$ & 49.9 & 53 & 51.4 \\
\footnotesize AMR + UD$^{++}$
& 74.9 & 72.7 & 73.8 & 47.1 & 50 & 48.5 \\
\footnotesize DM + UD$^{++}$
& 75.9$\star$ & 73.9$\star$ & \textbf{74.9}$\star$ & 48 & 54.8 & 51.2 \\
\footnotesize All
& 75.6$\star$ & 73.1 & 74.4$\star$ & 50.9 & 53.2 & 52
\end{tabular}
\caption{
Labeled precision, recall and $F_1$ (in~\%) for primary and remote edges,
on the \textbf{Wiki} test set.
$\star$~indicates significantly better than \textit{Single}.
HAR17: \citet{hershcovich2017a}.}\label{tab:id_results}
\end{table}

\begin{table}[t]
\centering
\small
\setlength\tabcolsep{3pt}
\begin{tabular}{l|lll|lll}
& \multicolumn{3}{c|}{\footnotesize \bf Primary} & \multicolumn{3}{c}{\footnotesize \bf Remote} \\
& \footnotesize \textbf{LP} & \footnotesize \textbf{LR} & \footnotesize \textbf{LF}
& \footnotesize \textbf{LP} & \footnotesize \textbf{LR} & \footnotesize \textbf{LF} \\
\hline
\multicolumn{4}{l|}{\small \bf English (out-of-domain)} & \\
\footnotesize HAR17
& 68.7 & 68.5 & 68.6 & 38.6 & 18.8 & 25.3 \\
\footnotesize Single
& 69 & 69 & 69 & 41.2 & 19.8 & 26.7 \\
\cline{1-1}
%\small \bf Multitask &&& \\
\footnotesize AMR
& 69.5 & 69.5 & 69.5 & 42.9 & 20.2 & 27.5 \\
\footnotesize DM
& 70.7$\star$ & 70.7$\star$ & 70.7$\star$ & 42.7 & 18.6 & 25.9 \\
\footnotesize UD$^{++}$
& 69.6 & 69.8$\star$ & 69.7 & 41.4 & 22 & 28.7 \\
\footnotesize AMR + DM
& 70.7$\star$ & 70.2$\star$ & 70.5$\star$ & 45.8 & 19.4 & 27.3 \\
\footnotesize AMR + UD$^{++}$
& 70.2$\star$ & 69.9$\star$ & 70$\star$ & 45.1 & 21.8 & 29.4 \\
\footnotesize DM + UD$^{++}$
& 70.8$\star$ & 70.3$\star$ & 70.6$\star$ & 41.6 & 21.6 & 28.4 \\
\footnotesize All
& 71.2$\star$ & 70.9$\star$ & \textbf{71}$\star$ & 45.1 & 22 & \textbf{29.6} \\
\hline
\multicolumn{4}{l|}{\small \bf French (in-domain)} & \\
\small Single & 68.2 & 67 & 67.6 & 26 & \enskip 9.4 & 13.9 \\
\small UD & 70.3 & 70$\star$ & \textbf{70.1}$\star$ & 43.8 & 13.2 & 20.3 \\
\hline
\multicolumn{4}{l|}{\small \bf German (in-domain)} & \\
\small Single & 73.3 & 71.7 & 72.5 & 57.1 & 17.7 & 27.1 \\
\small UD & 73.7$\star$ & 72.6$\star$ & \textbf{73.2}$\star$ & 61.8 & 24.9$\star$ & \textbf{35.5}$\star$
\end{tabular}
\caption{
Labeled precision, recall and $F_1$ (in~\%) for primary and remote edges,
on the \textbf{20K} test sets.
$\star$~indicates significantly better than \textit{Single}.
HAR17: \citet{hershcovich2017a}.}\label{tab:ood_results}
\end{table}

Table~\ref{tab:id_results} presents our results on the English in-domain Wiki test set.
MTL with all auxiliary tasks and their combinations improves the primary $F_1$ score over
the single task baseline. In most settings the improvement is statistically significant.
Using all auxiliary tasks contributed less than just DM and UD$^{++}$,
the combination of which yielded the best scores yet in in-domain UCCA parsing,
with 74.9\% $F_1$ on primary edges.
Remote $F_1$ is improved in some settings, but due to the relatively small number of remote
edges (about 2\% of all edges), none of the differences is significant.
Note that our baseline single-task model (\textit{Single})
is slightly better than the current state-of-the-art \cite[HAR17;][]{hershcovich2017a},
due to the incorporation of additional features (see \S\ref{sec:classifier}).

Table~\ref{tab:ood_results} presents our experimental results on the 20K corpora in the three languages.
For English out-of-domain, improvements from using MTL are even more marked. 
Moreover, the improvement is largely additive: the best model, using all three auxiliary tasks (\textit{All}),
yields an error reduction of 2.9\%. Again, the single-task baseline is slightly better than HAR17.

The contribution of MTL is also apparent in French and German in-domain parsing:
3.7\% error reduction in French
(having less than 10\% as much UCCA training data as English)
and 1\% in German, where the training set is comparable in size to the English one,
but is noisier (see~\S\ref{sec:experiments}).
The best MTL models are significantly better than single-task models,
demonstrating that even a small training set for the main task may suffice,
given enough auxiliary training data (as in French).

\section{Discussion}\label{sec:discussion}

\paragraph{Quantifying the similarity between tasks.}
Task similarity is an important factor in MTL success
\cite{E17-2026,E17-1005}.
In our case, the main and auxiliary tasks are annotated on different corpora
from different domains (\S\ref{sec:experiments}), and
the target representations vary both in form and in content.
%We found that the combination of many different domains therefore benefits domain
%adaptation even without auxiliary data from the target domain,
%in accordance with previous work \cite{Finkel2009JointPA}.

To quantify the domain differences, we follow \citet{Plank2011EffectiveMO} and measure the L1 distance 
between word distributions in the English training sets and 20K test set
(Table~\ref{tab:domain_sim}).
All auxiliary training sets are more similar to 20K than Wiki is, which may
contribute to the benefits observed on the English 20K test set.

\begin{table}[t]
\centering
\small
\begin{tabular}{l|cccc}
& \footnotesize 20K & \footnotesize AMR & \footnotesize DM & \footnotesize UD \\
\hline
\footnotesize Wiki & 1.047 & 0.895 & 0.913 & 0.843 \\
\footnotesize 20K && 0.949 & 0.971 & 0.904 \\
\footnotesize AMR &&& 0.757 & 0.469 \\
\footnotesize DM &&&& 0.754
\end{tabular}
\caption{L1 distance between dataset word distributions,
quantifying domain differences in English (low is similar).\label{tab:domain_sim}}
\end{table}

\begin{table}[t]
\centering
\small
\begin{tabular}{l|lll|lll}
& \multicolumn{3}{c|}{\footnotesize \bf Primary} & \multicolumn{3}{c}{\footnotesize \bf Remote} \\
& \footnotesize \textbf{UP} & \footnotesize \textbf{UR} & \footnotesize \textbf{UF}
& \footnotesize \textbf{UP} & \footnotesize \textbf{UR} & \footnotesize \textbf{UF} \\
\hline
AMR & 53.8 & 15.6 & 24.2 & \enskip 7.3 & \enskip 5.5 & \enskip 6.3 \\
DM & 65 & 49.2 & 56 & \enskip 7.4 & 65.9 & 13.3 \\
UD$^{++}$ & 82.7 & 84.6 & 83.6 & 12.5 & 12.7 & 12.6
\end{tabular}
\caption{Unlabeled $F_1$ scores between the representations of the same English sentences (from PTB WSJ), converted to the unified DAG format, and annotated UCCA graphs.\label{tab:common}}
\end{table}

As a measure of the formal similarity of the different schemes to UCCA,
we use unlabeled $F_1$ score evaluation on both primary and remote edges (ignoring edge labels).
To this end, we annotated 100 English sentences from Section 02 of the Penn Treebank Wall Street Journal
(PTB WSJ).
Annotation was carried out by a single expert UCCA annotator,
and is publicly available.\footnote{\url{http://github.com/danielhers/wsj}}
These sentences had already been annotated by the AMR, DM and PTB schemes,\footnote{We
convert the PTB format to UD$^{++}$ v1 using Stanford CoreNLP,
and then to UD v2 using Udapi: \url{http://github.com/udapi/udapi-python}.}
and we convert their annotation to the unified DAG format.

Unlabeled $F_1$ scores between the UCCA graphs and those converted from AMR, DM and UD$^{++}$
are presented in Table~\ref{tab:common}.
UD$^{++}$ is highly overlapping with UCCA, while DM less so, and AMR even less
(cf. Figure~\ref{fig:converted_examples}). 

Comparing the average improvements resulting from adding each of the tasks as auxiliary
(see \S\ref{sec:results}), we find 
AMR the least beneficial, UD$^{++}$ second, and DM the most beneficial,
in both in-domain and out-of-domain settings.
This trend is weakly correlated with the formal similarity between the tasks
(as expressed in Table~\ref{tab:common}), but weakly negatively correlated with the word distribution similarity scores (Table~\ref{tab:domain_sim}).
We conclude that other factors should be taken into account to fully explain this effect,
and propose to address this in future work through controlled experiments, where corpora of the same
domain are annotated with the various formalisms and used as training data for MTL.

%However, due to the deep model architecture (\S\ref{sec:multitask}),
%generalizations useful to all tasks (e.g., lexical) may be shared,
%even if not expressed directly in nodes and edges.
%These findings suggest the diversity of domains may have as important an effect on parsing performance, as the diversity of formalisms. 

\paragraph{AMR, SDP and UD parsing.} 
Evaluating the full MTL model (\textit{All}) on the unlabeled auxiliary tasks yielded
64.7\% unlabeled Smatch $F_1$ \cite{cai2013smatch} on the AMR development set,
when using oracle concept identification
(since the auxiliary model does not predict node labels),
27.2\% unlabeled $F_1$ on the DM development set,
and 4.9\% UAS on the UD development set.
These poor results reflect the fact that model selection was based on the score
on the UCCA development set, and that the model parameters dedicated to auxiliary tasks were
very limited (to encourage using the shared parameters).
However, preliminary experiments using our approach produced
promising results on each of the tasks' respective English development sets,
when treated as a single task:
67.1\% labeled Smatch $F_1$ on AMR
(adding a transition for implicit nodes and classifier for node labels),
79.1\% labeled $F_1$ on DM,
and 80.1\% LAS $F_1$ on UD.
For comparison, the best results on these datasets are 70.7\%, 91.2\% and 82.2\%, respectively
\cite{foland2017abstract,Peng-EtAl:2018:NAACL,K17-3002}.

\section{Conclusion}\label{sec:conclusion}

We demonstrate that semantic parsers can leverage a range of 
semantically and syntactically annotated data, to improve their performance.
Our experiments show that MTL improves UCCA parsing,
using AMR, DM and UD parsing as auxiliaries.
We propose a unified DAG representation, 
construct protocols for converting these schemes into the unified format,
and generalize a transition-based DAG parser to support all these tasks,
allowing it to be jointly trained on them.
%While the machine learning techniques we use are not novel,
%our contribution is in allowing these models to learn from
%significantly different tasks.

%A natural question is whether our method can benefit AMR, DM or UD parsing.
While we focus on UCCA in this work, our parser is capable of parsing any
scheme that can be represented in the unified DAG format,
and preliminary results on AMR, DM and UD are promising (see~\S\ref{sec:discussion}).
Future work will investigate whether a single
algorithm and architecture can be competitive on all of these parsing tasks,
an important step towards a joint many-task model for semantic parsing.

%%%%%%%%%%%%%%%%%%%%%%%%%%%%%%%%%%%%%%%%%%%%%%%%%%%%%%%%%%%%%%%
\section*{Acknowledgments}

This work was supported by the Israel Science Foundation (grant no. 929/17),
by the HUJI Cyber Security Research Center
in conjunction with the Israel National Cyber Bureau in the Prime Minister's Office,
and by the Intel Collaborative Research Institute for Computational Intelligence (ICRI-CI).
The first author was supported by a fellowship from the
Edmond and Lily Safra Center for Brain Sciences.
We thank Roi Reichart, Rotem Dror
and the anonymous reviewers for their helpful comments.

\bibliography{references}
\bibliographystyle{acl_natbib}

\end{document}

\maketitle

\paragraph{Features.}

Table~\ref{tab:features} lists all feature used for the classifier (see \S\ref{sec:classifier}).
Numeric features are taken as they are, whereas categorical features are mapped to real-valued embedding
vectors.
For \texttt{w} features,
we concatenate randomly-initialized and pre-trained word embeddings.
For each node, we select a \textit{head terminal} by traversing the graph according to
a priority order on edge labels, taken from \citet{hershcovich2017a}.

$s_i$ refers to stack node $i$ from the top, and
$b_i$ to buffer node $i$.
$xl$ and $xr$ refer to a $x$'s leftmost and rightmost children, and
$xL$ and $xR$ to its leftmost and rightmost parents.

\texttt{w} refers to the node's head terminal text,
\texttt{t} to its POS tag, and
\texttt{d} to its dependency relation.
\texttt{h} refers to the node's height,
\texttt{e} to the tag of its first incoming edge,
\texttt{n} and \texttt{c} to the node label and category (used only for AMR),
\texttt{p} to any separator punctuation between $s_0$ and $s_1$,
\texttt{q} to the count of any separator punctuation between $s_0$ and $s_1$,
\texttt{x} to the numeric value of gap type \cite{maier-lichte:2016:DiscoNLP},
\texttt{y} to the sum of gap lengths,
\texttt{P}, \texttt{C}, \texttt{I}, \texttt{E}, and \texttt{M} to the number of
parents, children, implicit children, remote children, and remote parents,
\texttt{N} to the numeric value of the head terminal's named entity IOB indicator,
\texttt{T} to its named entity type,
\texttt{\#} to its word shape (capturing orthographic features, e.g. "Xxxx" or "dd"),
\texttt{\^{}} to its one-character prefix, and
\texttt{\$} to its three-character suffix.

$x \to y$ refers to the existing edge from $x$ to $y$.
\texttt{x} is an indicator feature, taking the value of 1 if the edge exists or 0 otherwise,
\texttt{e} refers to the edge label, and
\texttt{d} to the dependency distance between the head terminals of the nodes.

$a_i$ to the transition taken $i+1$ steps ago.
\texttt{A} refers to the action type label (e.g. \textsc{shift}/\textsc{right-edge}/\textsc{node}), and
\texttt{e} to the edge label created by the action (e.g. $C$/$E$/$P$).

\texttt{node ratio} is the ratio between non-terminals and terminals, taken from \citet{hershcovich2017a}.

\begin{table}[h]
\centering
\begin{tabular}{l|l}
\bf Nodes & \bf Features \\ 
\hline
%       w   t   d   e   n   c   p   T   #   ^   $   x   h   q   y   P   C   I   E   M   N
$s_0$ & \texttt{wtdencpT\#\^{}\$xhqyPCIEMN} \\
$s_1$ & \texttt{wtdencT\#\^{}\$xhyN} \\
$s_2$ & \texttt{wtdencT\#\^{}\$xhy} \\
$s_3$ & \texttt{wtdencT\#\^{}\$xhyN} \\
$b_0$ & \texttt{wtdncT\#\^{}\$hPCIEMN} \\
$b_1, b_2, b_3$ & \texttt{wtdncT\#\^{}\$} \\
$s_0l, s_0r, s_1l, s_1r, s_0ll, s_0lr, s_0rl, s_0rr, s_1ll, s_1lr, s_1rl, s_1rr$ &
    \texttt{wenc\#\^{}\$} \\
$s_0L, s_0R, s_1L, s_1R, b_0L, b_0R$ & \texttt{wen\#\^{}\$} \\
\hline
\bf Edges & \\
$s_0 \to s_1, s_0 \to b_0$ & \texttt{xd} \\
$s_1 \to s_0, b_0 \to s_0$ & \texttt{x} \\
$s_0 \to b_0, b_0 \to s_0$ & \texttt{e} \\
\hline
\bf Past actions \\
$a_0, a_1$ & \texttt{eA} \\
\hline
\bf Misc. & \texttt{node ratio}
\end{tabular}
\caption{Transition classifier features.\label{tab:features}}
\end{table}

\paragraph{Conversion to and from Unified DAG Format.}

Although all experiments reported in the paper with the auxiliary tasks
(AMR, DM and UD) are using unlabeled parsing for these schemes,
our conversion code supports full conversion to and from these formats,
and is publicly available at \url{http://github.com/danielhers/semstr/tree/master/semstr/conversion}.

Conversion from AMR to the unifid DAG format and back
results in 95\% Smatch $F_1$ \cite{cai2013smatch} when averaged over the
LDC2017T10 test set.
On SDP, the conversion is lossless and results in identical graphs
when converted to UCCA and back.
For UD, and conversion results in 98.5\% LAS $F_1$ on the UD English test set,
due to multi-word tokens, not supported in the unified DAG format.

\paragraph{Qualitative evaluation.}

Figure~\ref{fig:qualitative} shows an example sentence from the English 20K test set,
with the outputs of both our single-task model and our best MTL model (using all auxiliaries).
While the single-task model obtains an $F_1$ score of 67.9\% on this sentence, the MTL model's output
matches the gold-annotates graph perfectly.
This example demonstrates how
the parser's ability to identify syntactic constituents,
which is important for all tasks we tackled, is improved with MTL.

\begin{sidewaysfigure}
    \centering
\scalebox{.6}{
\begin{tikzpicture}[->,level distance=26mm,
  level 1/.style={sibling distance=16cm},
  level 2/.style={sibling distance=7cm},
  level 3/.style={sibling distance=2cm},
  every circle node/.append style={fill=black}]
  \tikzstyle{word} = [font=\rmfamily,color=black]
  \node (1_1) [circle] {}
  {
  child {node (1_2) [circle] {}
    {
    child {node (1_5) [circle] {}
      {
      child {node (1_10) [word] {No} edge from parent node[auto] {\scriptsize $E$}}
      child {node (1_11) [word] {transoceanic} edge from parent node[auto] {\scriptsize $E$}}
      child {node (1_12) [word] {navigational} edge from parent node[auto] {\scriptsize $E$}}
      child {node (1_13) [word] {undertaking} edge from parent node[auto] {\scriptsize $C$}} }edge from parent node[auto] {\scriptsize $A$}}
    child {node (1_6) [circle] {}
      {
      child {node (1_14) [word] {has} edge from parent node[auto] {\scriptsize $F$}}
      child {node (1_15) [word] {been} edge from parent node[auto] {\scriptsize $E$}}
      child {node (1_16) [word] {conducted} edge from parent node[auto] {\scriptsize $C$}} }edge from parent node[auto] {\scriptsize $P$}}
    child {node (1_7) [circle] {}
      {
      child {node (1_17) [word] {with} edge from parent node[auto] {\scriptsize $R$}}
      child {node (1_18) [circle] {}
        {
        child {node (1_25) [circle] {}
          {
          child {node (1_28) [word] {more} edge from parent node[auto] {\scriptsize $E$}}
          child {node (1_29) [word] {ability} edge from parent node[auto] {\scriptsize $C$}} }edge from parent node[auto] {\scriptsize $C$}}
        child {node (1_26) [word] {,} edge from parent node[auto] {\scriptsize $U$}}
        child {node (1_27) [circle] {}
          {
          child {node (1_30) [word] {no} edge from parent node[auto] {\scriptsize $E$}}
          child {node (1_31) [circle] {}
            {
            child {node (1_32) [word] {business} edge from parent node[auto] {\scriptsize $E$}}
            child {node (1_33) [word] {dealings} edge from parent node[auto] {\scriptsize $C$}} }edge from parent node[auto] {\scriptsize $C$}} }edge from parent node[auto] {\scriptsize $C$}} }edge from parent node[auto] {\scriptsize $C$}} }edge from parent node[auto] {\scriptsize $A$}} }edge from parent node[auto] {\scriptsize $H$}}
  child {node (1_3) [circle] {}
    {
    child {node (1_8) [circle] {}
      {
      child {node (1_19) [word] {have} edge from parent node[auto] {\scriptsize $F$}}
      child {node (1_20) [word] {been} edge from parent node[auto] {\scriptsize $E$}}
      child {node (1_21) [word] {crowned} edge from parent node[auto] {\scriptsize $C$}} }edge from parent node[auto] {\scriptsize $P$}}
    child {node (1_9) [circle] {}
      {
      child {node (1_22) [word] {with} edge from parent node[auto] {\scriptsize $R$}}
      child {node (1_23) [word] {greater} edge from parent node[auto] {\scriptsize $D$}}
      child {node (1_24) [word] {success} edge from parent node[auto] {\scriptsize $P$}} }edge from parent node[auto] {\scriptsize $A$}} }edge from parent node[auto] {\scriptsize $H$}}
  child {node (1_4) [word] {.} edge from parent node[auto] {\scriptsize $U$}} }
;
\end{tikzpicture}}
\scalebox{.6}{
\begin{tikzpicture}[->,level distance=26mm,
      level 1/.style={sibling distance=10cm},
      level 2/.style={sibling distance=7cm},
      level 3/.style={sibling distance=2cm},
  every circle node/.append style={fill=black}]
  \tikzstyle{word} = [font=\rmfamily,color=black]
  \node (1_1) [circle] {}
  {
  child {node (1_2) [circle] {}
    {
    child {node (1_6) [circle] {}
      {
      child {node (1_12) [word] {No} edge from parent node[auto] {\scriptsize $E$}}
      child {node (1_13) [word] {transoceanic} edge from parent node[auto] {\scriptsize $E$}}
      child {node (1_14) [word] {navigational} edge from parent node[auto] {\scriptsize $E$}}
      child {node (1_15) [word] {undertaking} edge from parent node[auto] {\scriptsize $C$}} }edge from parent node[auto] {\scriptsize $A$}}
    child {node (1_7) [circle] {}
      {
      child {node (1_16) [word] {has} edge from parent node[auto] {\scriptsize $F$}}
      child {node (1_17) [word] {been} edge from parent node[auto] {\scriptsize $F$}}
      child {node (1_18) [word] {conducted} edge from parent node[auto] {\scriptsize $C$}} }edge from parent node[auto] {\scriptsize $P$}}
    child {node (1_8) [circle] {}
      {
      child {node (1_19) [word] {with} edge from parent node[auto] {\scriptsize $R$}}
      child {node (1_20) [word] {more} edge from parent node[auto] {\scriptsize $E$}}
      child {node (1_21) [word] {ability} edge from parent node[auto] {\scriptsize $C$}} }edge from parent node[auto] {\scriptsize $A$}} }edge from parent node[auto] {\scriptsize $H$}}
  child {node (1_3) [word] {,} edge from parent node[auto] {\scriptsize $U$}}
  child {node (1_4) [circle] {}
    {
    child {node (1_9) [circle] {}
      {
      child {node (1_22) [word] {no} edge from parent node[auto] {\scriptsize $E$}}
      child {node (1_23) [word] {business} edge from parent node[auto] {\scriptsize $E$}}
      child {node (1_24) [word] {dealings} edge from parent node[auto] {\scriptsize $C$}} }edge from parent node[auto] {\scriptsize $A$}}
    child {node (1_10) [circle] {}
      {
      child {node (1_25) [word] {have} edge from parent node[auto] {\scriptsize $F$}}
      child {node (1_26) [word] {been} edge from parent node[auto] {\scriptsize $F$}}
      child {node (1_27) [word] {crowned} edge from parent node[auto] {\scriptsize $C$}} }edge from parent node[auto] {\scriptsize $P$}}
    child {node (1_11) [circle] {}
      {
      child {node (1_28) [word] {with} edge from parent node[auto] {\scriptsize $R$}}
      child {node (1_29) [word] {greater} edge from parent node[auto] {\scriptsize $E$}}
      child {node (1_30) [word] {success} edge from parent node[auto] {\scriptsize $C$}} }edge from parent node[auto] {\scriptsize $A$}} }edge from parent node[auto] {\scriptsize $H$}}
  child {node (1_5) [word] {.} edge from parent node[auto] {\scriptsize $U$}} }
;
\end{tikzpicture}}
\caption{Output of single-task model on sentence 53001 from the English 20K test set (top),
and of MTL model using all of AMR, DM and UD$^{++}$ as auxiliaries on the same sentence (bottom).
\label{fig:qualitative}}
\end{sidewaysfigure}

\bibliography{references}
\bibliographystyle{acl_natbib}